\documentclass{article}

\usepackage[final,nonatbib]{neurips_2023}

\usepackage[utf8]{inputenc} 
\usepackage[T1]{fontenc}    
\usepackage{hyperref}       
\usepackage{url}            
\usepackage{booktabs}       
\usepackage{amsfonts}       
\usepackage{nicefrac}       
\usepackage{microtype}      
\usepackage{xcolor}         
\usepackage{graphicx}
\usepackage{subcaption}
\usepackage{tabularx}       
\usepackage[square,sort,comma,numbers]{natbib}

\title{Exploring the Hyperparameter Space of Image Diffusion Models for Echocardiogram Generation}

\author{%
  Hadrien~Reynaud\\
  Imperial College London, London, UK \\
  FAU Erlangen-Nürnberg, DE \\
  \texttt{hadrien.reynaud19@imperial.ac.uk} \\
  \And
  Bernhard~Kainz\\
  Imperial College London, London, UK \\
  FAU Erlangen-Nürnberg, DE \\
  \texttt{bernhard.kainz@fau.de} 
}

\begin{document}
\maketitle

\begin{abstract}
This work presents an extensive hyperparameter search on Image Diffusion Models for Echocardiogram generation. The objective is to establish foundational benchmarks and provide guidelines within the realm of ultrasound image and video generation. This study builds over the latest advancements, including cutting-edge model architectures and training methodologies. We also examine the distribution shift between real and generated samples and consider potential solutions, crucial to train efficient models on generated data. We determine an Optimal FID score of $0.88$ for our research problem and achieve an FID of $2.60$. This work is aimed at contributing valuable insights and serving as a reference for further developments in the specialized field of ultrasound image and video generation.
\end{abstract}

\section{Introduction}
Ultrasound imaging is a pivotal component in medical diagnostics, offering a non-invasive approach to visualize internal structures and functions of the human body. As it serves as a primary technology for screening and diagnosing various conditions, there's a need to advance our analysis capabilities of ultrasound images. One avenue to do so is through the generation of specific echocardiogram images and videos, representing edges cases, and the generation of large scale, balanced datasets.

Training clinicians on unusual/edge cases in ultrasound is essential for fostering better diagnostic capabilities in ambiguous situations. Furthermore, creating synthetic ultrasound data should lead to a wider distribution of such data, while maintaining patient privacy and enabling collaborative advancements.

In this study, we establish state-of-the-art baselines for ultrasound image generation, by conducting hyperparameter searches on diffusion models. While there has been notable progress in this domain, \cite{reynaud2023feature,reynaud_dartagnan_2022,salehi_patient-specific_2015}, there is still a lack of exhaustive exploration of the solutions available for generating ultrasound data.

\section{Method}

We train image diffusion models using three different architectures: the Ablated Diffusion Model (ADM)~\cite{dhariwal2021diffusion}, the Denoising Diffusion Probabilistic Models ++ (DDPM++)~\cite{song2021scorebased} and the Noise Conditional Score Network ++ (NCSN++)~\cite{song2021scorebased}. All models use the Elucidated Diffusion Model (EDM)~\cite{karras2022elucidating} sampler for training, as it has been shown~\cite{balaji2022ediffi,kim2023refining} to yield the best FID scores.

After establishing which is the best architecture, we explore the distribution shift between real and generated samples by training classifiers. This distribution shift prevents models trained on generated data to perform well on real data. To solve this problem, we adopt an idea similar to Discriminator Guidance (DG)~\cite{kim2023refining}, an extension of the classifier guidance paradigm focusing on solving the distribution shift at little cost and with no major drawback. 
As a last verification step, we train classifiers to appreciate the improvement brought by DG to the sample's quality.

\section{Experimental Settings}
We focus on echocardiogram images, with the intent of setting a baseline for image generation, which can be extended to the much more demanding task of echocardiogram video generation \cite{reynaud2023feature,reynaud_dartagnan_2022}. 
For all presented experiments, the Fréchet inception distance (FID)  score~\cite{heusel2018gans} is computed over 50,000 samples.
We use the Echonet-Dynamic dataset \cite{ouyang_video-based_2020}, which consists of 10,030 echocardiogram videos of resolution $112\times112$. 
We extract 1-in-5 frames to generate a dataset of 358,259 images, which are resized to $64\times64$ to drastically speed up training and generation times.

We go through a series of experiments to determine which is the best FID score we can obtain on this $64\times64$.

We compute "Optimal FIDs" for both dataset, which act as baselines, and give us a trend on the impact of resolution over the FID. The Optimal FID is computed by splitting the real data collection in two, and then computing the FID between these splits. This yields an Optimal FID of $0.95$ for the $112\times112$ resolution and $0.88$ for the $64\times64$ resolution.

\section{Results}
We follow an iterative process of evaluating all available options and keeping only the best one to move onto the next hyperparameter search.

\textbf{Best Architecture} We train all the architectures for 24 hours on nodes of 8 NVIDIA A40, with 5\% dropout and no data augmentation, as altering the real echocardiograms colorimetry and structure would degrade the performance. The architectures are parametrized with their default configuration, as defined by their respective authors. This also defines their number of parameters.

\begin{figure}[h]
    \centering
    \begin{minipage}{.55\textwidth}
        \centering
        \resizebox{\textwidth}{!}{
        \begin{tabular}{lcccccccc}
        \toprule
        Model   & Sampler & Batch & K. img & M. Param & FID $\downarrow$ \\
        \midrule
        ADM~\cite{dhariwal2021diffusion} & EDM \cite{karras2022elucidating} & 512  & 30,106 & 295.1 & 3.75 \\ 
        DDPM++~\cite{song2021scorebased} & EDM \cite{karras2022elucidating} & 512 & 60,211 & 55.7  & 5.67 \\ 
        NSCN++~\cite{song2021scorebased} & EDM \cite{karras2022elucidating} & 512  & 60,211 & 56.4  & 5.24 \\ 
        \midrule 
        ADM~\cite{dhariwal2021diffusion} & VP \cite{song2021scorebased}     & 512  & 30,106 & 295.1 & 23.96 \\ 
        ADM~\cite{dhariwal2021diffusion} & VE \cite{song2021scorebased}     & 512  & 30,106 & 295.1 & 588.521 \\ 
        \midrule 
        DDPM++~\cite{song2021scorebased} & VP \cite{song2021scorebased}     & 512  & 60,211 & 55.7 & 286.726 \\ 
        NSCN++~\cite{song2021scorebased} & VE \cite{song2021scorebased}     & 512  & 60,211 & 56.4 & 425.325 \\ 
        \bottomrule
        \end{tabular}%
        }
        \captionof{table}{Architecture hyperparamater search. K.~img is the number of images seen during training in thousands. M.~Param is the number of model's parameters in millions.}
        \label{tab:exp_setup}
    \end{minipage}%
    \hfill
    \begin{minipage}{.45\textwidth}
        \centering
        \includegraphics[width=\linewidth]{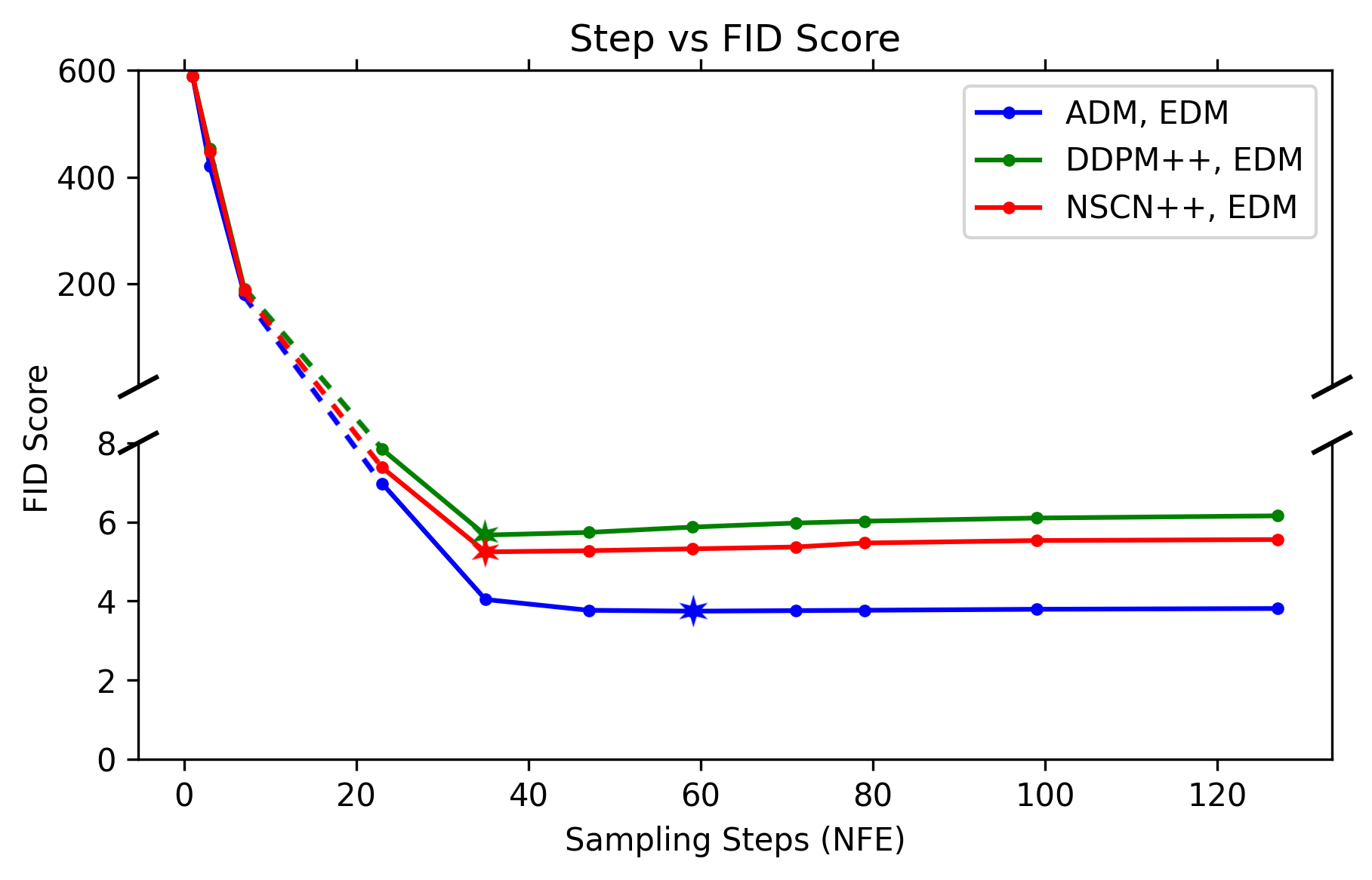}
        \caption{FID per NFE.}
        \label{fig:fid_step}
    \end{minipage}
\end{figure}

In Table~\ref{tab:exp_setup}, the best performing model is the ADM-EDM~\cite{dhariwal2021diffusion}, with an FID of 3.75 when generating over 59 Neural Function Executions (NFE), equivalent to 30 EDM steps. We can see on Figure~\ref{fig:fid_step} that the FID decreases abruptly once we reach the 35 NFE mark, and plateaus afterward. 

It is important to note that the slight variation we observe are not due to a margin of error. We demonstrate this by running 2 experiments, one where we fix the generated samples and vary the real samples 10 times, and a second one where we fix the real samples and generate 10 times 50,000 samples. In the case of the ADM-EDM sampled over 18 EDM steps, we observe an FID of $4.02\pm0.045$ in the first case and $4.05\pm0.001$ in the second case. As only the real dataset subsample can vary the score, we run all our experiments over the same subsample, leading to highly accurate FID scores.

In Table~\ref{tab:exp_setup}, results for the Variance Preserving (VP) and Variance Exploding (VE) setups are also shown. Training models with these setups lead to drastically poorer performance, which means that they are not as versatile as the EDM. Further experiments would be required to find the exact hyperparameters required to make these combinations of model and architecture work for echocardiograms, but we decided to keep the original setups for this study.

\textbf{Discriminator Guidance} Given the large FID difference between our optimal FID (0.88) and our current best FID (3.75), we assume a distribution shift, which we quantify in the next section. We attempt to reduce it by training a classifier to discriminate between the real and generated samples, and propagating the information learned by the classifier into the reverse diffusion process, as detailed by~\cite{kim2023refining}. 
We train a UNet-encoder to discriminate between real and generated samples, of our ADM-EDM, at all noise levels. To do so, we partially apply the forward noising process over both sets of images. 

Once our discriminator is trained, we plug it in the reverse diffusion process as described in~\cite{kim2023refining}, and samples 50,000 for every epoch, in order to find the best performing one. We find that, for the best epoch with the nest set of hyperparameters, the FID improves from 3.75 to 2.60, showing the significance of DG. We use 30 (59 NFE) sampling steps, a weight of 5 for the first order EDM prediction, no guidance for the correction (weight 0) and a scaling of 2 for the Discriminator Guidance vectors.

\textbf{Distribution Shifts} In order to quantitatively appreciate the distribution shift in the generated samples, we train two classifiers: (A) a simple linear classifier with no non-linearity to detect pixel-level shift, and (B) a resnet-18 to grasp the key differences in higher-complexity structures. We train these classifiers before and after applying DG, leading to the results presented in Table~\ref{tab:shift}. The classifiers are trained over 50 epochs with a learning rate of 1e-4, binary cross entropy loss and Adam optimizer. We use 50,000 real and 50,000 generated samples with a 90/10 split for the training/validation set.

\begin{figure}[h]
    \centering
    \begin{minipage}{.6\textwidth}
        \centering
        \resizebox{\textwidth}{!}{
            \begin{tabular}{lccc}
            \toprule
            Classifier  & Linear    & ResNet-18 & Aug. ResNet-18\\
            \midrule
            Pre-DG      & 71.67\%   & 66.25\%   & 98.30\%  \\
            Post-DG     & 72.14\%   & 60.91\%   & 84.24\%  \\
            \bottomrule
            \end{tabular}
        }
        \captionof{table}{Classifiers accuracy at discriminating real vs generated samples. Aug. ResNet-18 is the resnet-18 architecture with resize, crop, flip, and rotation augmentation.}
        \label{tab:shift}
    \end{minipage}%
    \hfill
    \begin{minipage}{.35\textwidth}
        \centering
        \setcounter{figure}{2} 
        \begin{subfigure}{.48\linewidth}
            \centering
            \includegraphics[width=\linewidth]{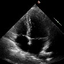}
            \subcaption{Real}
            \label{subfig:a}
        \end{subfigure}%
        \hfill
        \begin{subfigure}{.48\linewidth}
            \centering
            \includegraphics[width=\linewidth]{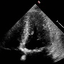}
            \subcaption{Generated}
            \label{subfig:b}
        \end{subfigure}
        \setcounter{figure}{1} 
        \caption{Echocardiogram samples.}
        \label{fig:pair}
    \end{minipage}
\end{figure}

While DG does improve the FID of our samples, according to Table~\ref{tab:shift}  it has no impact on how easy it is for a neural network to differentiate between real and generated samples. This means that when training a downstream task on generated images, the DG samples will still yield downgraded performance, at least because of the distribution shift in the inputs at training vs. inference time. We also observe in Table~\ref{tab:shift} that the linear classifier performs better than ResNet-18, which is due to ResNet-18 overfitting to the train set. We corrected this by using non-destructive data augmentation, leading to the augmented (Aug.) ResNet-18 results.

We show a real Figure~\ref{subfig:a} and ADM-EDM-DG-generated echocardiogram sample in Figure~\ref{subfig:b}. Due to the complexity of echocardiogram images compared to natural images, it is extremely difficult for humans to distinguish between real and generated ones. It was shown in~\cite{reynaud2023feature} that even experts can't detect generated echocardiograms, which makes the process of improving the generation process all the more challenging.


\section{Discussion \& Conclusion}
We conduct a large hyperparameter search over the current most widely spread diffusion model architectures (ADM, DDPM++, NSCN++) and training setups (EDM, VE, VP). We show that the ADM architecture, which has more parameters by default, yields better results. 
We also show that EDM is the best framework for the task at hand, largely beating VE and VP setups.
Through experimentation, we learn that Discriminator Guidance does improve the FID score of our model. However, it does not help with the distribution shift, and it remains easy for a neural network, even a simple one, to discriminate between real and generated samples.

In the future, we intend to explore Discriminator Guidance further, by conducting additional hyperparameters search on Discriminator Guidance only, including switching the discriminator architecture and exploring different gradient weighting mechanisms.


\newpage


\section*{Acknowledgements}

This work was supported by Ultromics Ltd., the UKRI Centre for Doctoral Training in Artificial Intelligence for Healthcare (EP/S023283/1) and HPC resources provided by the Erlangen National High Performance Computing Center (NHR@FAU) of the Friedrich-Alexander-Universität Erlangen-Nürnberg (FAU) under the NHR project b143dc and b180dc. NHR funding is provided by federal and Bavarian state authorities. NHR@FAU hardware is partially funded by the German Research Foundation (DFG) – 440719683. Support was also received from the ERC - project MIA-NORMAL 101083647 and DFG KA 5801/2-1, INST 90/1351-1.

\section*{Potential Negative Societal Impacts}
Publicly releasing generative models instead of actual datasets in medical imaging research is an emerging idea which call for new ethical guidelines and discussions. These models can create medical images that, while not sourced from real patients, maintain the essential clinical features required for research. This approach could circumvent the cumbersome ethical approval processes associated with using real patient data, enabling a more efficient research cycle and the wider dissemination of highly protected medical data. It offers a privacy-centric alternative to releasing whole datasets, allowing medical researchers to access necessary resources without violating privacy rights, potentially accelerating breakthroughs in medical imaging, thereby enhancing healthcare outcomes.

While the application of generative models promises advancements in medical imaging research, it inevitably prompts substantial deliberations regarding their ethical deployment and regulation. Ensuring the accuracy and unbiased nature of the synthesized images is paramount to uphold the sanctity and integrity of medical research. Achieving equilibrium between groundbreaking innovation and stringent ethical norms is indispensable, necessitating the creation of comprehensive regulatory frameworks. These frameworks must enforce transparency, accountability, and strong adherence to the key principles of medical ethics. The inception of generative models as alternatives to actual datasets could be a major step forward, but their application mandates meticulous evaluation and substantial supervision to focus their capabilities in advancing medical imaging research.

A third aspect to account for is the inherent capability of generative models to learn, and possibly regenerate, specific patient information. While these models are designed to create synthetic, non-specific patient images, the depth of learning could lead to the generation of images bearing significant resemblance to the original data, potentially exposing sensitive patient information. This possibility raises additional ethical considerations about the extent of information that a model learns and retains, and whether it could inadvertently reveal identifiable patient information. Addressing these concerns is crucial in fostering an environment of trust and safety around the use of generative models, ensuring the ethical boundaries are respected while continuing to innovate in the domain of medical imaging research.

\bibliographystyle{plain}
\bibliography{bib.bib}

\begin{thebibliography}{10}

\bibitem{balaji2022ediffi}
Yogesh Balaji, Seungjun Nah, Xun Huang, Arash Vahdat, Jiaming Song, Karsten Kreis, Miika Aittala, Timo Aila, Samuli Laine, Bryan Catanzaro, et~al.
\newblock ediffi: Text-to-image diffusion models with an ensemble of expert denoisers.
\newblock {\em arXiv preprint arXiv:2211.01324}, 2022.

\bibitem{dhariwal2021diffusion}
Prafulla Dhariwal and Alex Nichol.
\newblock Diffusion models beat gans on image synthesis, 2021.

\bibitem{heusel2018gans}
Martin Heusel, Hubert Ramsauer, Thomas Unterthiner, Bernhard Nessler, and Sepp Hochreiter.
\newblock Gans trained by a two time-scale update rule converge to a local nash equilibrium, 2018.

\bibitem{karras2022elucidating}
Tero Karras, Miika Aittala, Timo Aila, and Samuli Laine.
\newblock Elucidating the design space of diffusion-based generative models, 2022.

\bibitem{kim2023refining}
Dongjun Kim, Yeongmin Kim, Se~Jung Kwon, Wanmo Kang, and Il-Chul Moon.
\newblock Refining generative process with discriminator guidance in score-based diffusion models, 2023.

\bibitem{ouyang_video-based_2020}
David Ouyang, Bryan He, Amirata Ghorbani, Neal Yuan, Joseph Ebinger, and et~al. Langlotz, Curtis~P.
\newblock Video-based {AI} for beat-to-beat assessment of cardiac function.
\newblock {\em Nature}, 580:252--256, April 2020.

\bibitem{reynaud2023feature}
Hadrien Reynaud, Mengyun Qiao, Mischa Dombrowski, Thomas Day, Reza Razavi, Alberto Gomez, Paul Leeson, and Bernhard Kainz.
\newblock Feature-conditioned cascaded video diffusion models for precise echocardiogram synthesis.
\newblock {\em arXiv preprint arXiv:2303.12644}, 2023.

\bibitem{reynaud_dartagnan_2022}
Hadrien Reynaud, Athanasios Vlontzos, Mischa Dombrowski, Ciarán Gilligan~Lee, Arian Beqiri, and et~al. Leeson, Paul.
\newblock D’{ARTAGNAN}: {Counterfactual} {Video} {Generation}.
\newblock In {\em MICCAI}, pages 599--609, 2022.

\bibitem{salehi_patient-specific_2015}
Mehrdad Salehi, Seyed-Ahmad Ahmadi, Raphael Prevost, Nassir Navab, and Wolfgang Wein.
\newblock Patient-specific {3D} {Ultrasound} {Simulation} {Based} on {Convolutional} {Ray}-tracing and {Appearance} {Optimization}.
\newblock In {\em MICCAI}, pages 510--518, 2015.

\bibitem{song2021scorebased}
Yang Song, Jascha Sohl-Dickstein, Diederik~P. Kingma, Abhishek Kumar, Stefano Ermon, and Ben Poole.
\newblock Score-based generative modeling through stochastic differential equations, 2021.

\end{thebibliography}

\end{document}